\documentclass[10pt, draftclsnofoot, twocolumn]{ieee-sty/IEEEtran}

\usepackage[acronym]{glossaries}
\usepackage[ruled,linesnumbered]{algorithm2e}
\usepackage{dutchcal}
\usepackage{amsmath}
\usepackage{enumerate}
\usepackage{graphicx}
\usepackage{float} 
\usepackage{subfigure}
\usepackage{tabularx}
\usepackage{booktabs}
\usepackage{amssymb}
\usepackage{cite}
\usepackage{multirow}
\usepackage{multicol}
\usepackage{siunitx}
\usepackage{footmisc}
\usepackage{amsmath}
\usepackage{setspace}

\singlespace

\makeglossaries

\newacronym{IEEE}{IEEE}{Institute of Electrical and Electronics Engineers}

\begin{document}


\def \ArticleTitle{ESFL: Efficient Split Federated Learning over Resource-Constrained Heterogeneous Wireless Devices}

\def \AuthorA{Guangyu Zhu, Yiqin Deng,~\IEEEmembership{Member,~IEEE}, Xianhao Chen,~\IEEEmembership{Member,~IEEE},  Haixia Zhang,~\IEEEmembership{Senior Member,~IEEE}, Yuguang Fang,~\IEEEmembership{Fellow,~IEEE}, Tan F. Wong,~\IEEEmembership{Member,~IEEE}}

\def \footnoteA{ \thanks{Guangyu Zhu and Tan F. Wong are with Department of Electrical and Computer Engineering, University of Florida, Gainesville, FL 32611, USA. (e-mail: gzhu@ufl.edu, twong@ece.ufl.edu).}

\thanks{Yiqin Deng and Haixia Zhang are with School of Control Science and Engineering and with Shandong Key Laboratory of Wireless Communication Technologies, Shandong University, Jinan 250061, Shandong, China (e-mail: yiqin.deng@email.sdu.edu.cn; haixia.zhang@sdu.edu.cn).}

\thanks{Xianhao Chen is with  Department of Electrical and Electronic Engineering, University of Hong Kong, Pok Fu Lam, Hong Kong, China (e-mail: xchen@eee.hku.hk).}

\thanks{Yuguang Fang is with Department of Computer Science, City University of Hong Kong, Kowloon, Hong Kong, China (e-mail: my.fang@cityu.edu.hk).}

\thanks{This work was supported in part by US National Science Foundation under grant CNS-2106589.}

\thanks{Corresponding author: Yiqin Deng.}

}


\newcommand {\Title} {\ArticleTitle}

\newcommand {\Authors} {\IEEEauthorblockN{\AuthorA}}

\newcommand\blfoottext[1]{%
    \bgroup
    \renewcommand\thefootnote{\fnsymbol{footnote}}%
    \renewcommand\thempfootnote{\fnsymbol{mpfootnote}}%
    \footnotetext[0]{#1}%
    \egroup
}

\def \hasBibliography{1}

\title{\Title{\footnoteA}}

\author{\Authors}
\maketitle

\newcommand{\etal}{\textit{et al}.}
\newcommand{\ie}{\textit{i}.\textit{e}.}
\newcommand{\eg}{\textit{e}.\textit{g}.}


\begin{abstract}
    Federated learning (FL) allows multiple parties (distributed devices) to train a machine learning model without sharing raw data. How to effectively and efficiently utilize the resources on devices and the central server is a highly interesting yet challenging problem. In this paper, we propose an efficient split federated learning algorithm (ESFL) to take full advantage of the powerful computing capabilities at a central server under a split federated learning framework with heterogeneous end devices (EDs). By splitting the model into different submodels between the server and EDs, our approach jointly optimizes user-side workload and server-side computing resource allocation by considering users' heterogeneity. We formulate the whole optimization problem as a mixed-integer non-linear program, which is an NP-hard problem, and develop an iterative approach to obtain an approximate solution efficiently. Extensive simulations have been conducted to validate the significantly increased efficiency of our ESFL approach compared with standard federated learning, split learning, and splitfed learning.  
\end{abstract}


\begin{IEEEkeywords}
    Distributed machine learning, federated learning, split learning, wireless networking.
\end{IEEEkeywords}



%


\section{Introduction}
\label{sections:introduction}
In recent years, machine learning (ML) has attracted intensive attention in many fields and numerous artificial intelligence (AI) applications, such as computer vision, smart health, connected and autonomous driving, information access control, and security surveillance~\cite{khan2021federated}. According to Cisco~\cite{nguyen2021federated}, there were nearly $850$ zettabyte of data generated by people, machines, and things at the network edge in $2021$. It is definitely infeasible to simply send this huge data volume to a central server to process or compute. To do so, a tremendous network bandwidth is required, incurring intolerable latency. Hence, distributed machine learning algorithms have been developed to cope with the aforementioned challenges for large geographically distributed volume of data by distributing the ML workloads to EDs~\cite{qiu2016survey}. Moreover, awareness and concerns about users' privacy have been raised in the digitalized world~\cite{zhang2021survey}. Following the Privacy-by-Design (PbD) principle, the best way to achieve user privacy is not to disclose raw data, and so it would be more effective for EDs not to share private raw data as much as possible during the machine learning process.


\textit{Federated learning} (FL), a privacy-preserving distributed ML technique enables  EDs to collaboratively learn a global ML model without sharing their raw data, consequently reducing the requirement of communication bandwidth between EDs and a central server as well. The intuitive idea of privacy-preserving distributed ML was investigated independently by some researchers Xu et al. \cite{xu2014control, xu2015privacy, xu2017my}, Shokri et al.~\cite{shokri2015privacy}, Gong et al. \cite{gong2015privacy}. McMahan et al.~\cite{mcmahan2017communication} first constructed a distributed ML framework based on decentralized datasets held by different users, and coined their algorithm as FL. The original FL algorithm updates ML models on EDs locally, and aggregates the updated ML models to obtain the global model at a central server remotely, in contrast to the traditional ML requiring EDs to upload their private data to where the ML model is trained. Since FL retains users' private data on EDs without sharing raw data with the server, the server only aggregates users' locally updated models, and thus the privacy of end users is naturally preserved  to some extent. Besides, as all training processes are performed by the EDs, the computation and communication capacities of EDs can also be utilized. FL provides various advantages, such as data privacy preservation, reduced communication latency, and enhanced learning performance~\cite{nguyen2021federated}. However, pushing all training workload to EDs sometimes is impracticable. Training complex ML models often consumes unacceptable amount of training memory and computing power on Internet of Things (IoT) devices with limited communication and computing resources, and incurs intolerable latency. Besides, the ML model size sometimes reaches GBs and even TBs(e.g., GPT-1 has 0.12 billion parameters, GPT-2 has 1.5 billion parameters, whereas GPT-3 has more than 175 billion parameters), resulting in a significant communications burden duo to the need of frequently uploading local ML models.

\begin{figure}[!htbp]
    \centering
    \includegraphics[width=0.45\textwidth]{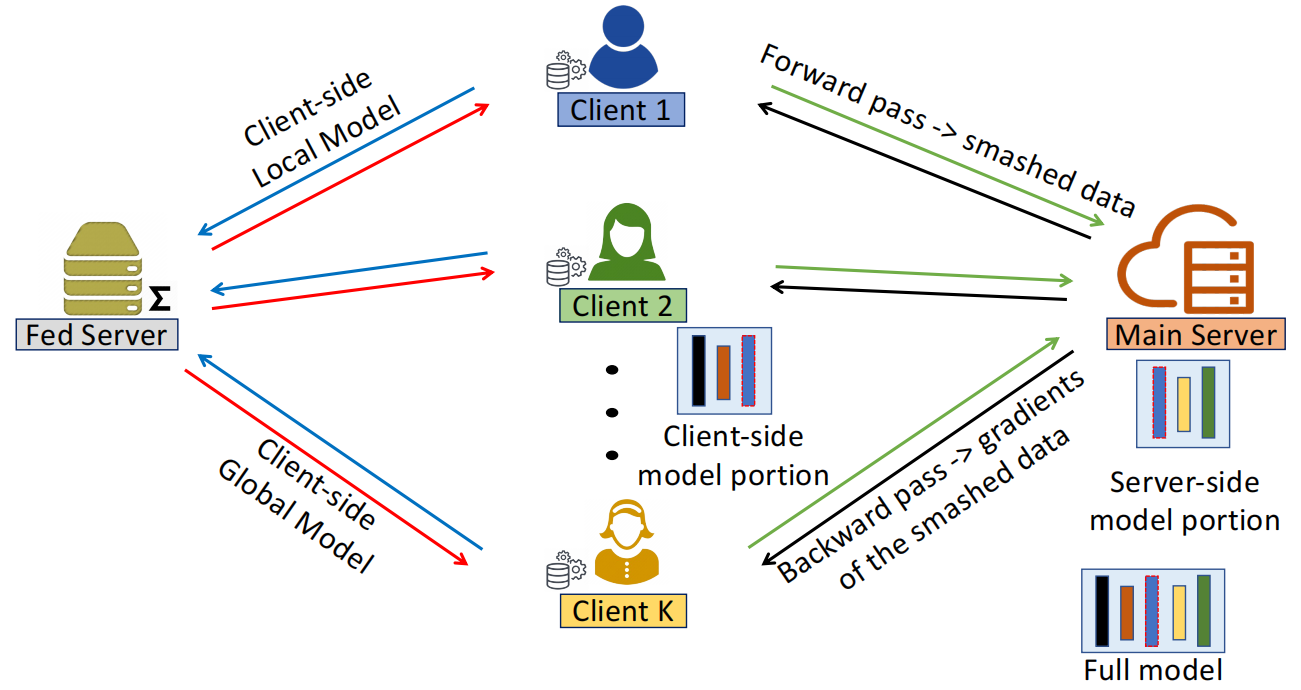}
    \caption{The architecture of splitfed learning (SFL) system.}
    \label{splitfed}
\end{figure}

To cope with the dilemma between insufficient ED resources and complicated ML models, we leverage another ML technique, called split federated learning (SFL)~\cite{thapa2022splitfed}, which introduces model splitting from split learning (SL)~\cite{vepakomma2018split} to FL. The SFL framework was proposed by Thapa et al.~\cite{thapa2022splitfed} in 2020 to integrate federated learning with split learning, shown in Figure~\ref{splitfed}. Under SFL framework, the main server helps the training process for each client, and the Fed server is tasked with the aggregation of all locally updated models. The SFL approach presents a compelling advantage for resource-constrained environments, since the main server undertake parts of local training workload.

From the conventional FL and SFL implementations across heterogeneous devices, we have identified several limitations that can significantly impede the system's overall performance and efficiency. In synchronous FL and SFL, the aggregation process is inherently constrained by the pace of the slowest participant due to the necessity for the server to collect updated local gradients from all selected EDs. This synchronicity results in a scenario where devices with abundant computational resources are rendered idle as they wait for the transmission of local models from less capable devices, often referred to as ``stragglers''. This inefficiency is primarily attributed to the varying and unpredictable communication and computational capabilities of heterogeneous EDs. Therefore, addressing resource heterogeneity (RH), also called the system heterogeneity in FL, is critical to enhancing the efficiency and overall feasibility of SFL.



In this paper, we propose a novel \textit{efficient split federated learning} algorithm (ESFL) to boost training efficiency and performance by considering the privacy-preserving constraints. The cornerstone of ESFL lies in its proactive strategic utilization of heterogeneity in system resources and device capabilities. Unlike standard synchronous FL and SFL, our approach involves the server in the training phase but dynamically adjusts the distribution of user-side workload and server-side resources. The ESFL algorithm thereby capitalizes on the intrinsic resource variability across EDs to optimize ML outcomes and system efficiency.

Our major contributions in this paper are summarized as follows.
\begin{itemize}
    \item [1)] We design ESFL, a novel distributed machine learning framework, which significantly improves the training efficiency of SFL by taking ED heterogeneity into consideration.
    \item [2)] We formulate a mixed-integer non-linear program (MINLP) by jointly considering the allocation of user-side workload (model separation) and server-side resource, and develop an iterative optimization algorithm to find a suboptimal solution with a low time complexity.
    \item [3)] We evaluate the performance of our ESFL approach through extensive simulations compared to the state-of-the-art methods such as FL, SL and SFL, and demonstrate the superiority of the proposed ESFL framework.
\end{itemize}

The remainder of this paper is organized as follows. In Section~\ref{sec:RelateWork}, we discuss related works about FL, SFL, and RH. In Section~\ref{sec:ESFLearning}, we expound on the framework and system model of ESFL. In Section~\ref{sec:Optimization}, we present the optimization problem formulation and the solution to the joint resource allocation and model separation problem. In Section~\ref{sec:Experiments}, results of extensive simulations and experiments are presented. In Section~\ref{sec:conclusion}, we summarize the proposed algorithm and experimental results and conclude the paper.


\section{Related Works}
\label{sec:RelateWork}
There exist quite extensive research on splitting learning (SL) and federated learning (FL) in the current literature. In this section, we will concentrate on closely related works to review. In~\cite{ang2020robust}, Ang et al. offered a robust architecture for FL to increase communication efficiency by reducing transmission noise in wireless networks. To address the communications overhead in FL, Wang et al.~\cite{wang2019cmfl} developed an algorithm, named Communication Mitigated Federated Learning (CMFL), to eliminate irrelevant local model updates that were trained over users' biased data. It should be noted that the aforementioned two methods only address communication efficiency in FL, as all training processes are executed by users at EDs. FedMMD was proposed in~\cite{yao2018two} to reduce communication and computing costs during the local training process in FL using a two-stream model with Maximum Mean Discrepancy (MMD) to replace the local training for a single model in FedAVG~\cite{mcmahan2017communication}. However, while this method reduces the number of communication rounds, it concurrently increases the user-side workload due to the need of computing the MMD loss functions. Shi et al.~\cite{shi2020joint} address both communication and computing resource heterogeneities in wireless FL by providing a joint device scheduling and resource allocation strategy. Despite this integrated approach, the limitation of FL is still evident, as the user-side workload cannot be reduced. There exist also some other innovative approaches to ease the work load for communications and computing for federated learning. Watanabe et al. \cite{watanabe2023novel} and Chen et al. \cite{chen2022federated} leveraged wireless mesh networks to either facilitate communications or reduce communications traffic. Guo et al. \cite{guo2024federated, guo2023federated} utilized edge nodes and federated reinforcement learning to help resource-constrained D2D devices in industrial IoT systems and 5G networks, which can address the device heterogeneity issue to some extent.   


In~\cite{thapa2022splitfed}, Thapa et al. designed a novel framework, called \textit{split federated learning} (SFL), to take advantage of parallel training among different users in FL and the model splitting in SL to reduce the computing workload on EDs. In~\cite{gao2020end}, Gao et al. implemented SFL and evaluated the performance on IoT devices. In \cite{lin2023efficient}, Lin et al. developed an efficient parallel split learning algorithm by applying the last-layer gradient aggregation to reduce communication and computing overheads in SL. In \cite{wu2022split}, Wu et al. designed a resource allocation strategy for cluster-based SFL. In \cite{kim2022bargaining}, Kim et al. devised a bargaining game to negotiate the cut layer in personalized parallel SL. All aforementioned literature on SFL overlook the crucial aspect of the joint consideration of both user-side resource and workload heterogeneities. This oversight is evidenced by the fact that the cut layer remains the same at all EDs. However, it is obvious that adjusting cut layers can significantly change user-side workload distribution.

To reduce the computing and communication workload at EDs when considering RH in FL, Sattler et al. ~\cite{sattler2019robust} designed a novel model compression algorithm to extend the commonly used top-k gradient sparsification to FL to compress both model uploading and downloading. However, since model compression inevitably results in performance degradation, it is advisable to maintain the integrity of the model architecture throughout the training process. With the emergence of edge computing~\cite{deng2022actions}, utilizing the computing resources at the both the central server and edges can be leveraged to reduce workload at EDs. For instance, in~\cite{wang2019edge}, Wang et al. proposed to enable EDs to collaborate with edge nodes by exchanging model parameters to reduce the user-side workload, and to apply deep reinforcement learning to optimize the operations of multi-access edge computing (MEC), caching, and communications. Several researchers attempted to leverage edge nodes to assist FL by fully uploading local training tasks to trusted edge nodes to reduce the user-side workloads~\cite{lim2020federated}. Unfortunately, the trustworthiness of training edge nodes is often hard to guarantee. Our ESFL algorithm introduce an integrated strategy for the allocation of server-side computing resource and user-side training workloads. This approach is designed to accommodate the variations in computing and communication resources inherent in heterogeneous EDs. We will present the detailed design next. 
\section{Efficient Split Federated Learning}
\label{sec:ESFLearning}
\subsection{Motivation}
While SFL only address the resource constraints at EDs, it does not account for the variations in data distribution (DH) and resource availability (resource heterogeneity or RH) inherent in EDs. Within the SFL framework, clients or EDs are required to partition their models following an identical structure, leading to a scenario where the communication and computational workload on the client side is influenced solely by the volume of data. In contrast, our Efficient Split Federated Learning (ESFL) takes a holistic approach, considering both the client-side workload and the server-side computing resource allocation, which is designed to mitigate ``stragglers'' problem in FL, thereby enhancing the training efficiency of SFL. It is worth noting that our method is a scheme for joint optimization of resource and workload allocation, which can be integrated with any user selection algorithms presently existed in FL. The ESFL framework demonstrates the capability to increase training efficiency across the board, independent of the particular user selection algorithm integrated in the framework. This underscores the inherent adaptability and effectiveness of our proposed scheme in enhancing the training efficacy of FL.
\subsection{ESFL framework}
In this subsection, we elaborate our framework of ESFL consist of four components, namely, \textit{split training}, \textit{federated aggregation}, \textit{communication model}, and \textit{resource allocation}. We assume that an ML task is composed of multiple training rounds. The FL server first initializes ML model in the initial round of training and the subsequent training rounds. The one-round training procedure is given as follows.

\begin{itemize}
    \item [1)] \textbf{User selection}: The server selects users randomly from the \textit{available users} (EDs), who are willing and ready to participate this round of training (these users are called the \textit{selected users}).\;
    \item [2)] \textbf{Model splitting and resource allocation}: The server first acquires the information of the selected EDs, including data amount, available communication, and computing resources. Then, the server jointly splits the ML model into user-side models and server-side models (the so-called \textit{cut layer decision}) and allocates adequate server-side resources based on the users' information and the ML model structure for the \textit{selected user}.\;
    \item [3)] \textbf{Model distribution}: The server distributes user-side models (with the corresponding architectures) to the \textit{selected users}.\;
    \item [4)] \textbf{Split training}: The server and all \textit{selected users} then collaboratively update both user-side and server-side ML models simultaneously and the server determines the ML hyperparameters, such as learning rate, data batch size, and local training epochs.\;
    \item [5)] \textbf{Federated aggregation}: After repeating several epochs of the \textbf{split training} step, all selected users that have finished their training upload their updated user-side ML models to the server, and then the server generates an updated global ML model based on the user-side ML models collected from the \textit{selected users} and the corresponding server-side ML models at the server.
\end{itemize}

\begin{table}[!htbp]
    \centering
    \caption{Summary of Important Notations of ESFL}
    \scriptsize{6}{6}
    \begin{tabularx}{0.45\textwidth}{p{0.06\textwidth}X}
        \toprule
        \textbf{Notation} & \textbf{Description}\\
        \midrule
        $S$ & The set of selected users\\
        $R$ & The total training rounds\\
        $\epsilon_i$ & The number of local training epochs of user $i$\\
        $\textbf{W}_{u, i}^{r, e}$ & The user-side model for user $i$ at epoch $e$ at round $r$\\
        $\textbf{W}_{s, i}^{r, e}$ & The server-side model for user $i$ at epoch $e$ at round $r$\\
        $\textbf{A}_{l_i}^{r, e}$ & The activation calculated by the user $i$ at epoch $e$ at round $r$ using forward propagation algorithm\\
        $\eta$ & The step size factor of the federated aggregation\\
        $N$ & The number of total one-round training data samples\\
        $n_i$ & The number of one-round training data samples for user $i$\\
        $\textbf{W}^{r}$ & The global model at round $r$\\
        $\textbf{W}_i^{r}$ & The cascaded local model generated by user $\textbf{W}_{u, i}$ and server-side model $\textbf{W}_{s, i}$ at round $r$\\
        $b_{i}^u$ & Uploading data rate of user $i$\\
        $b_{i}^d$ & Downloading data rate of user $i$\\
        $B_i$ & The bandwidth allocated to user $i$\\
        $P^u_{i}$ & The uplink transmission power of user $i$\\
        $P^d_{i}$ & The downlink transmission power of user $i$\\
        $\gamma^u_{i}$ & The uplink channel gain of user $i$\\
        $\gamma^d_{i}$ & The downlink channel gain of user $i$\\
        $N_{0}$ & The noise power density\\
        $T$ & The total training time\\
        $T_{r, i}$ & The $r$-th round training time for user $i$\\
        $T_{i}^U$ & The $r$-th round local model uploading time for user $i$\\
        $T_{i}^D$ & The $r$-th round global model downloading time for user $i$\\
        $T_{i}^e$ & The split training time at epoch $e$ for user $i$\\
        $T^{agg}$ & The federated aggregation time for the server at round $r$\\
        $M^{l_i}_{i}$ & The local model uploading/ downloading time for user $i$\\
        $t_i^{e, c}$ & The user-side computing time for user $i$ at epoch $e$\\
        $t_i^{e, b}$ & The uploading time of the user-side activation for user $i$ at epoch $e$\\
        $t_i^{e, C}$ & The server-side computing time for user $i$ at epoch $e$\\
        $t_i^{e, B}$ & The downloading time of the server-side updated activation for user $i$ at epoch $e$\\
        $D_c^l$ & The user-side computing workload of the cut layer $l$ for training one sample\\
        $D_b^l$ & The user-side communication workload of the cut layer $l$ for training one sample\\
        $D$ & The computing workload for training one data sample\\
        $c_i$ & The local available computing resource for user $i$\\
        $C_i$ & The allocated server-side computing resource to user $i$\\
        $x_i$ & The cut layer indicator vector of user $i$\\
        $s_i$ & The available storage space for user $i$\\
        $m_i$ & The available memory space for user $i$\\ 
        \bottomrule
    \end{tabularx}
    \label{Notation_ESFL}
\end{table}

In what follows, we will provide more details for the whole procedure below. 

\subsubsection{Split Training}
Different from federated learning, which only relies on EDs to update local ML models, we split the local training across EDs and the server. Similar to the \textit{SplitFed}~\cite{thapa2022splitfed}, local training is repeatedly conducted $\epsilon_i$ epochs for user $i$ before one-round global ML model aggregation at the server. The user-side ML model for user $i$ is $\mathcal{\textbf{W}}_{u, i}^{r, e}$ at epoch $e$ in the $r$-th round. We denote the updated local user-side ML model at the epoch $e+1$ as

\begin{equation}
    \begin{aligned}
        \mathcal{\textbf{W}}_{u, i}^{r, e+1} &= BP(\mathcal{\textbf{W}}_{u, i}^{r, e}, \rho_r, \nabla{\textbf{A}}_{l_i}^{r, e}),
    \end{aligned}
\end{equation}
where BP is the backpropagation algorithm, $\rho_r$ is the local learning rate in the $r$-th round, and $\nabla{\textbf{A}}_{l_i}^{r, e} = {\textbf{A}}_{l_i}^{r, e} - \hat{\textbf{A}}_{l_i}^{r, e}$ is the activation difference, which is the loss value for BP, and $\hat{\textbf{A}}_{l_i}^{r, e}$ is the updated activation calculated by the server using backpropagation algorithm with the shared labels and the estimated labels. 

\begin{figure}[!htbp]
    \centering
    \includegraphics[width=0.3\textwidth]{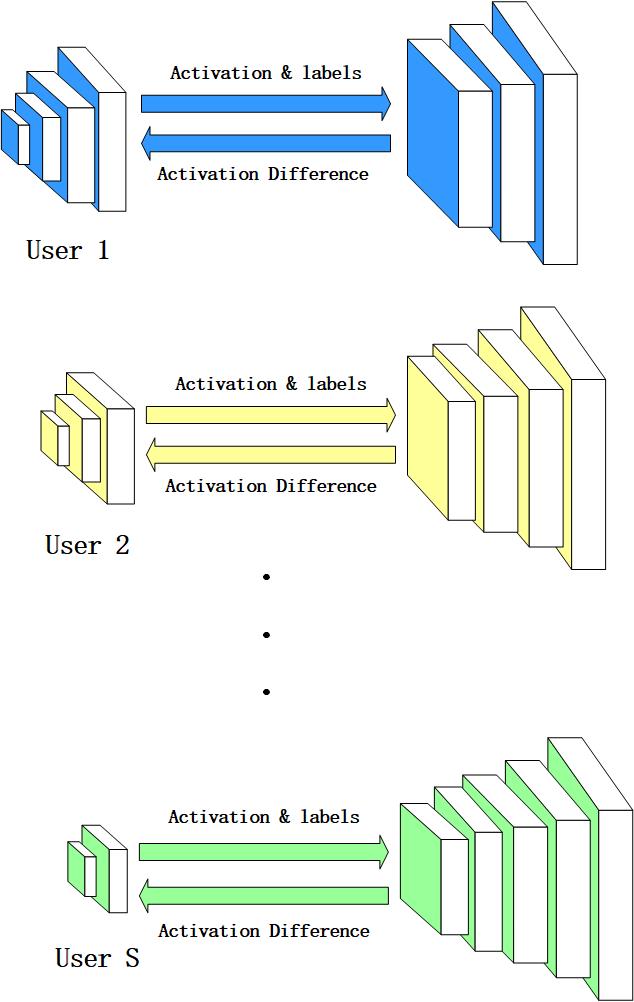}
    \caption{The split training procedure, where all selected users simultaneously transmit activation data and labels to the server, and the server sends back the corresponding activation difference.}
    \label{Split_training}
\end{figure}

The server-side ML model for user $i$ at epoch $e+1$ in the $r$-th round is given by 

\begin{equation}
    \begin{aligned}
        \mathcal{\textbf{W}}_{s, i}^{r, e+1} &= BP(\mathcal{\textbf{W}}_{s, i}^{r, e}, \rho_r, \nabla\mathcal{l}(\mathcal{\textbf{W}}_{s, i}^{r, e}, \hat{\textbf{Y}},\textbf{Y})).
    \end{aligned}
\end{equation}
To update the server-side ML model, we use the loss function $\mathcal{l}(\mathcal{\textbf{W}}_{s, i}^{r, e}, \hat{\textbf{Y}},\textbf{Y})$,  where $\hat{\textbf{Y}}$ is the estimated result computed by the server-side ML model and $\textbf{Y}$ is the true label shared by the user. The estimated result $\hat{\textbf{Y}}$ is the output of the current server-side ML model $\mathcal{\textbf{W}}_{s, i}^{r, e}$ and the input data ${\textbf{A}}_{l_i}^{r, e}$ is the activation generated by the user $i$'s samples and user-side ML model. Split training divides the local training into four parts in FL, \textit{i.e.}, user-side forward propagation, server-side forward propagation, server-side backpropagation, and user-side backpropagation. The entire split training procedure is shown in Figure~\ref{Split_training}.

The pseudocode for our ESFL is shown in Algorithm~\ref{SplitUpdate_code} and Algorithm~\ref{ESFL_code}. The split training is shown in Algorithm~\ref{SplitUpdate_code}, while Step $2$ is to let users send activations of the cut layer together with the label to the server.

\begin{algorithm*}
    \SetAlgoNoLine
    \SetKwInput{Input}{Input}
    \SetKwInOut{Output}{Output}
    \caption{SplitUpdate}
    \label{SplitUpdate_code}
    \Input{At epoch $e$ of round $r$, user-side model $\mathcal{\textbf{W}}_{u, i}^{r, e}$, server-side model $\mathcal{\textbf{W}}_{s, i}^{r, e}$, and local learning rate $\rho$\;
    }
    \Output{Updated user-side $\mathcal{\textbf{W}}_{u, i}^{r+1, e}$ and server-side model $\mathcal{\textbf{W}}_{s, i}^{r+1, e}$\;
    }
    User $i$ forwards propagation of $\mathcal{\textbf{W}}_{u, i}^{r, e}$ and generate activation $\textbf{A}_{l_i}^{r, e}$ and the label vector $\textbf{Y}$\;
    User $i$ sends $\textbf{A}_{l_i}^{r, e}$ at the cut layer and $\textbf{Y}$ to the server\;
    The server uses backpropagation to calculate the server-side gradient $\nabla\mathcal{l}(\mathcal{\textbf{W}}_{s, i}^{r, e}, \textbf{Y})$ and the activation difference $\nabla{\textbf{A}}_{l_i}^{r, e}$\;
    The server-side model update: $\mathcal{\textbf{W}}_{s, i}^{r, e+1} = BP(\mathcal{\textbf{W}}_{s, i}^{r, e}, \rho, \nabla\mathcal{l}(\mathcal{\textbf{W}}_{s, i}^{r, e}, \textbf{Y}))$\;
    The server sends $\nabla{\textbf{A}}_{l_i}^{r, e}$ to user $i$\;
    The user-side model update: $\mathcal{\textbf{W}}_{u, i}^{r, e+1} = BP(\mathcal{\textbf{W}}_{u, i}^{r, e}, \rho, \nabla{\textbf{A}}_{l_i}^{r, e})$
\end{algorithm*}

\subsubsection{Federated Aggregation}
\begin{figure}[h]
    \centering
    \includegraphics[width=0.45\textwidth]{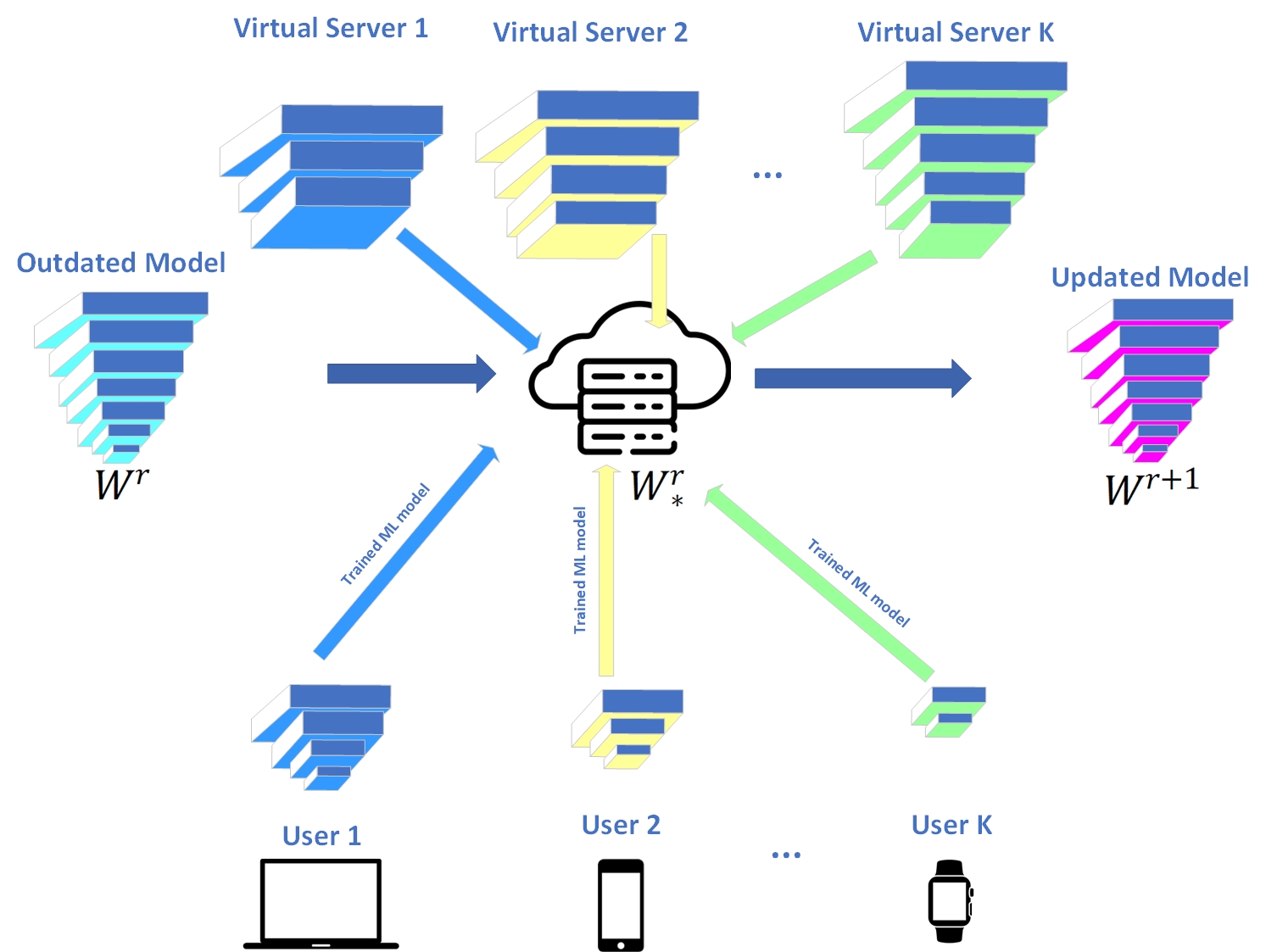}
    \caption{This figure illustrates the federated aggregation, where the previous global ML model is $\mathcal{\textbf{W}}^{r}$, the aggregated global ML model is $\mathcal{\textbf{W}}^{r}_{*}$ and the updated global ML model is $\mathcal{\textbf{W}}^{r+1}$}
    \label{Federated_aggregation}
\end{figure}

In ESFL, we leverage the \textsl{FedAVG} algorithm \cite{mcmahan2017communication} to aggregate multiple user-side and server-side updated ML models using $\eta$ as a step size factor. We call this aggregation method {\em federated aggregation}, which mitigates training oscillations by leveraging the global ML model in the previous round. This differs from FedAVG which only aggregates the current-round updated local ML models to generate the next-round global ML model. The architecture of federated aggregation is shown in Figure~\ref{Federated_aggregation}. We define the whole ML model trained by user $i$ in round $r$ as $\mathcal{w}_{i}^r$ and the global model at round $r$ is $\mathcal{\textbf{W}}^r$. The update of the global ML model at round ${r+1}$ is

\begin{center}
\begin{equation}
    \begin{gathered}
        \mathcal{\textbf{W}}^{r + 1} = \mathcal{\textbf{W}}^r - \eta * \left(\mathcal{\textbf{W}}^r - \mathcal{\textbf{W}}^r_{*}\right) \\
        = \mathcal{\textbf{W}}^r - \eta * \left(\mathcal{\textbf{W}}^r - \sum_{i} \frac{n_i*\mathcal{\textbf{W}}_i^{r}}{N}\right), 
    \end{gathered}
\end{equation}    
\end{center}
where $N = \sum_{i} n_i$ indicating the number of total training samples. 
        
The whole training model is split into two parts, namely, user-side model $\textbf{W}_{u,i}^r$ trained by a user-side ED and the server-side model $\textbf{W}_{s,i}^r$ trained by a virtual server $v_i$. All virtual servers are either virtual machines or containers at the central server, which are allocated with corresponding computing and communication resources according to the model training demands at this round. Thus, after concatenating the user-side ML model and the server-side ML model, for each selected user, the concatenated ML model will have the same architecture as the global ML model. The updated concatenated ML model for user $i$ at round $r + 1$ is
\begin{center}
    \begin{equation}
        \begin{aligned}
            \mathcal{\textbf{W}}^{r + 1}_i \gets \{\mathcal{\textbf{W}}^{r + 1}_{u,i}, \mathcal{\textbf{W}}^{r + 1}_{s,i}\}.
        \end{aligned}
    \end{equation}
\end{center}

In Algorithm~\ref{ESFL_code}, the resource allocation scheme and cut layer decision based on idle resource states of selected users are shown in Step $5$. In this section, we focus on synchronous federated learning, where the aggregation of all selected user-side ML models and server-side models can only be conducted after one-round training is finished.

\begin{algorithm*}[h]
    \caption{Efficient Split Federated Learning}
    \label{ESFL_code}
    \SetAlgoNoLine
    \SetKwInput{Input}{Input}
    \SetKwInOut{Output}{Output}
    \Input{The number of model aggregation round $R$, global learning rate $\eta$, local learning rate $\rho$, and user $i$'s number of local epochs $\epsilon_i$\;}
    \Output{Final Updated global model $\textbf{W}^{R+1}$\;}

    Initialize the global model parameters $\mathcal{\textbf{W}}$\;
    \For{$r = 1, 2, ..., R$}{
        The server randomly selects users $S$ to participate the $r$-th round training\;
        The server acquires states of available computing resource $c^*$, communication resource (uplink data rate) $b^*$, memory $m^*$, and storage space $s^*$ from $S$\;
        The server allocates its resources $C^*$ and $B^*$ and determines cut layer $l^*$ based on $c^*$ and $b^*$\;
        \For{selected user $i = 1, 2, ..., |S|$}{
            Send $\mathcal{\textbf{W}}_{u, i}^r$ to user $i$ based on $l_i$ and $\mathcal{\textbf{W}}^{r-1}$\;
            \For{local epoch $e = 1, 2, ..., \epsilon_i$}{
                $\mathcal{\textbf{W}}_{u, i}^{r, e+1}, \mathcal{\textbf{W}}_{s, i}^{r, e+1}$$\gets$\textbf{SplitUpdate}($\mathcal{\textbf{W}}_{u, i}^{r, e},\mathcal{\textbf{W}}_{s, i}^{r, e},\rho$)\;
            }
            User $i$ sends back the updated $\mathcal{\textbf{W}}_{u, i}^{r}$ to the server\;
        }
        Server-side model update: $\mathcal{\textbf{W}}_{s}^{r+1} = \mathcal{\textbf{W}}_{s}^{r} - \eta\sum_i \frac{n_i}{N}\nabla\mathcal{\textbf{W}}_{s, i}^{r}$\;
        User-side model update: $\mathcal{\textbf{W}}_{u}^{r+1} = \mathcal{\textbf{W}}_{u}^{r} - \eta\sum_i \frac{n_i}{N}\nabla\mathcal{\textbf{W}}_{u, i}^{r}$\;
    }
\end{algorithm*}

\subsubsection{Communication Model}
Since we only intend to demonstrate the effectiveness of our proposed ML scheme, we will adopt a simple communication system model for our study. For uplink and downlink transmissions, we assume uploading and downloading transmission bandwidth are equal. Specifically, the orthogonal multiple access (OMA) techniques are adopted where each user can be allocated with one orthogonal spectrum band for the needed data transmissions determined by the server (\textit{e.g.}, the base station). The uploading and downloading data rate for user $i$ is given by:
\begin{equation}
    \begin{aligned}
        b_{i}^u &=  B_{i}\log \left(1+\frac{P^u_{i} \gamma_{i}^u}{B_{i}N_{0}}\right),\\ 
        b_{i}^d &=  B_{i}\log \left(1+\frac{ P^d_{i} \gamma_{i}^d}{B_{i}N_{0}}\right),
    \end{aligned}
    \label{communication_equation}
\end{equation}
where the communication bandwidth allocated to the user $i$ is $B_i$, and the total available bandwidth is $B$ where $\sum_{i=1}^S B_i \leq B$, $\gamma_{i}^u$ is the uplink channel gain and $\gamma_i^d$ is the downlink channel gain for user $i$, $P^u_{i}$ and $P^d_{i}$ are the uplink and downlink transmission powers, respectively, when the resource block $B_{i}$ is used, where $P^u_{i}$, $P^d_{i}$, $\gamma_{i}^u$, and $\gamma_{i}^d$ are predetermined constant values, and $N_0$ is the noise power density. We assume that the transmission environment is stationary during one training round. For example, EDs can be cameras in smart homes, whose deployment locations remain fixed for a relatively long period.

\subsubsection{Workload and Resource Allocation}
To reduce the training workload on user-side EDs, we split ML training into local training and server ML training. Thus, the next problem is how to appropriately determine user-side communication and computing workload, alongside the strategic allocation of computing resources on the server to different virtual servers. Due to the varying sizes of data collected by different users/EDs and the heterogeneous available resources on EDs, simply assigning the same amount of resources and randomly choosing the cut layers for different users will not help the training time and the training performance. To maximize the training efficiency, one should address a joint optimization of workload and resource.

We denote the total training time as $T$ and the $r$-th round training time as $T_r$. To cope with the aforementioned joint optimization problem, we formulate the total training time as 
\begin{equation}
    \begin{gathered}
        T = \sum_{r = 1}^R T_r = \sum_{r = 1}^R \max_{i} T_{r, i},
    \end{gathered}
\end{equation}
where $T_{r, i}$ is the $r$-th round training time for user $i$. 

Since the times of individual training rounds are independent, minimizing the total training time is equivalent to minimizing the training time for each training round. Thus, in the subsequent development, we will only focus on one round of training and omit the training round index r for notational simplicity. One-round training time is composed of four parts, namely, model distribution time $T^{D}_{i}$, model upload time $T^{U}_{i}$, model aggregation time $T^{agg}$, and training time $T^e_{i}$ for local epoch $e$. Thus, the one-round training time can be expressed as
\begin{equation}
    \begin{gathered}
        T_{i} = T^{U}_{i} + T^{D}_{i} + {\sum_{e = 1}^{\epsilon_i} T^{e}_{i}} + T^{agg},\\
        T_{i}^{U} = \frac{M^{l_i}_{i}}{b_{i}^u},~T_{i}^{D} = \frac{M^{l_i}_{i}}{b_{i}^d},\\
    \end{gathered}
\end{equation}
where $b_{i}^U$ and $b_{i}^D$ are the uplink and downlink data rate for user $i$ as defined in (\ref{communication_equation}). We denote the index of the cut layer for user $i$ as $l_i$ and $\epsilon_i$ as the number of local epochs for user $i$. The uploading and downloading workload are assumed to be $M^l_{i}$, equal to the user-side model size of user $i$. Since each selected user applies the same local dataset to train ML model and the idle resource is also stationary, to simplify the following optimization problem, in this paper, we assume that the training times of all epochs are constant in each round of training time.

Since both user-side EDs and the server participate in each epoch model updating, one-epoch training time contains four parts, namely, local/user-side computing time $t^{e, c}$, uploading time for activation $t^{e, b}$, remote/server-side computing time $t^{e, C}$, and downloading time for updated activation $t^{e, B}$. Thus, one epoch time can be denoted as 
\begin{equation}\label{Epoch_time} 
    \begin{gathered}
        T^{e}_i = t^{e, c}_i + t^{e, b}_i + t^{e, C}_i + t^{e, B}_i \\
        = \frac{w^{e, c}_i}{c_i} + \frac{w^{e, b}_i}{b_i} + \frac{W^{e, C}_i}{C_i} + \frac{w^{e, B}_i}{B_i} \\
        = \frac{\sum_{l} {D^l_cx^l_i \cdot n_i}}{c_i} + \frac{\sum_{l} {D^l_bx^l_i \cdot n_i}}{b^u_i} + \\
        \frac{ (D - \sum_{l} {D^l_cx^l_i }) \cdot n_i}{C_i} + \frac{\sum_{l} {D^l_bx^l_i} \cdot n_i}{b^d_i}.\\
    \end{gathered}
\end{equation}

The cut layer for user $i$ is $l_i \in [1, 2, ..., L]$, where the whole ML neural network is composed of $L$ layers, and the server-side model and the user-side model for user $i$ are split at the $l_i$-th layer. To simplify the notation of workloads, we denote $x^l_i$ as an indicator function, where $\sum_l x^l_i = 1$, and $x_i^l=1$ indicates that $l$ layer is chosen as the cut layer for user $i$  while $x_i^l=0$ indicates that $l$ layer is not selected as the cut layer. The total computing workload for training one sample is $D$ and the user-side computing workload for user $i$ is $w^c_{i} = \sum_{l} {D^l_cx^l_i \cdot n_i}$, where $D^l$ is the computing workload for one training sample for user-side EDs when the cut layer is $l$. The upload and download data size for user $i$ is $w_{i}^b = \sum_{l} {{D^l_bx^l_i} \cdot n_i}$, where $D^l_b$ is the size of activation data for one training sample while the cut layer is $l$. For user $i$, computing capability is $c_i$, uplink transmission rate is $b^u_i$, and downlink transmission rate is $b^d_i$. The computing resource that the server allocates to user $i$ is $C_i$.

\section{Optimization and Solution Approach}
\label{sec:Optimization}
The ultimate objective of the ESFL training is to minimize the total training time. However, since every round training time are independent, minimizing the total training time is equivalent to minimizing the total training time in each round including computing and communication time given by (as we mentioned earlier, we will omit the dependence of the round number $r$ for notational simplicity) 
\begin{equation}
    \begin{gathered}
        \min T = \min \sum_r \max_{i} T_{i}.
    \end{gathered}
\end{equation}
To run Algorithm 2, we need to consider a few optimization problems for resource allocation, which should be solved during the ESFL training. The joint resource allocation and model splitting in our ESFL is a min-max optimization problem. To linearize the formulated optimization problem, we introduce an auxiliary variable $T_{max}$, which is no less than the training time for the straggler (\textit{i.e.}, the client who takes the longest time to complete one-round training). Thus, our problem can be formulated as
\begin{equation}\label{joint_optimize}
    \begin{gathered}
        \min_{{x_i^l}, C_i} T_{max}\\
        s.t.~T_{i}\leq T_{max},~i\in \{1,...,S\}\\
        \sum_{i=1}^S C_i \leq C_{total},\\
        \sum_{l=1}^{L}{x_i^l}M^{l}_{i} \leq s_i,\\
        \sum_{l=1}^{L}{x_i^l}m^{l}_{i} \leq m_i,\\
        ~\sum_{l=1}^{L} x^l_i = 1, ~x^l_i \in \{0, 1\}.\\
    \end{gathered}
\end{equation}
The total computing resource owned by the server is $C_{total}$, $M^{l}_{i}$ is the data size of user $i$'s user-side ML model, and $s_i$ is the available storage space at user $i$. To compute the user-side model, $m^{l}_{i}$ is the required memory space, and $m_i$ is the available memory space for user $i$. We denote $x_i = \{x_i^1, ..., x_i^L\}$ as the cut layer indicator vector, where $x^l_i$ indicates whether the ML model is split at layer $l$, in the sense that $x^l_i = 1$ when $l = l_i$, and $x^l_i = 0$ otherwise.

\begin{algorithm}[h]
    \SetAlgoNoLine
    \SetKwInput{Input}{Input}
    \SetKwInOut{Output}{Output}
    \caption{Alternative Optimization}
    \label{Alternative_optimization}
    \textbf{Initialization:} Allocating identical computing resource to all users, $C_i = \frac{C_{total}}{|S|}$\;
    \While{$C_i^{n-1} = C_i^{n}$}{
        Obtain the optimal $\{l_*\}$ of subproblem for cut layer decision for given $\{C_*\}$ by solving (\ref{layer_decision})\;
        Obtain the optimal $\{C_*\}$ of subproblem for resource allocation for the given cut layer decision $\{l_*\}$ by solving (\ref{resource_allocation})\;}
    \Output{$\{l_*\}$, $\{C_*\}$ for problem (\ref{joint_optimize})}
\end{algorithm}

The alternative optimization algorithm is shown in Algorithm~\ref{Alternative_optimization}. For the notational convenience, we use $\{C_*\}$ to denote $\{C_1, C_2, ..., C_S\}$ and $\{l_*\}$ denote $\{l_1, l_2, ..., l_S\}$. The reason why we use an alternative optimization approach is that the cut layer decisions for different selected users lead to the variance at the user-side workload. Moreover, the allocation of server-side computing resource to individual one user will affect the availability of resources for others, given the fixed total capacity of server-side resources, where the joint optimization of worklaods and resources incurs the coupling effect for different users. To address this coupling effect, our ESFL algorithm transforms the optimization problem into a mixed-integer non-linear program (MINLP), which is typically NP-hard. To solve the problem efficiently, we decompose it into two subproblems and solve them iteratively. We construct the first subproblem for cut layer decision by treating computing resource allocation as fixed decision variables:
\begin{equation}\label{layer_decision}
    \begin{gathered}
        \min_{x_i^l} T^e_i = \frac{\sum_{l} {D^l_cx^l_i \cdot n_i}}{c_i} + \frac{\sum_{l} {D^l_bx^l_i \cdot n_i}}{b^U_i} + \\
        \frac{ (D - \sum_{l} {D^l_cx^l_i }) \cdot n_i}{C_i} + \frac{\sum_{l} {D^l_bx^l_i} \cdot n_i}{b^D_i}\\
        s.t.~\sum_{l=1}^{L} x^l_i = 1, ~x^l_i \in \{0, 1\},\\
        M^{l}_{i} \leq s_i,\\
        m^{l}_{i} \leq m_i.\\
    \end{gathered}
\end{equation}
Leveraging the iterative optimization approach, the correlation between different users can be eliminated in the sense that we can focus on solving the first subproblem for each user independently, as shown in (\ref{layer_decision}). This is because the cut layer decision for each user is independent of others when computing resource allocation is given. The resulting subproblem for cut layer decision can be easily solved by a linear programming (LP) solver or exhaustive search with the time complexity reduced from $\mathcal O(L^S)$ to $\mathcal O(SL)$.

Based on the determined cut layers, we construct the second subproblem for the resource allocation scheme for computing resources for user $i$ as
\begin{equation}\label{resource_allocation}
    \begin{gathered}
        \min_{C_i} \max T^e_i = \frac{w_c^{l_i} \cdot n_i}{c_i} + \frac{w_b^{l_i} \cdot n_i}{b^u_i} + \\\frac{ (D - w_c^{l_i}) \cdot n_i}{C_i} + \frac{w_b^{l_i} \cdot n_i}{b^d_i}\\
        s.t.~\sum_{i=0}^S 0 \leq C_i \leq C_{total},
    \end{gathered}
\end{equation}
where communication and computing workloads for all users are constant since the cut layers $l_i$ are predetermined by solving the previous subproblem. Plus, the downlink $b^d_i$, uplink communication resource $b^u_i$ and user-side available computing resource $c_i$ are constant. Therefore, the equation (\ref{resource_allocation}) can be abbreviated as:
\begin{equation}\label{min_max_1}
    \begin{gathered}
        \min_{C_i} \max T^e_i =  \frac{a}{C_i} + b,
    \end{gathered}
\end{equation}
where $a$ and $b$ are constant. To solve this min-max problem, we assume there exists a variable $K$, where for all $C_i$, $K \geq \frac{a}{C_i} + b$. Then, we construct the equation(\ref{min_max_1}) as a minimizing problem:
\begin{equation}\label{min_max_2}
    \begin{gathered}
        \min K\\
        s.t.~\sum_{i=0}^S 0 \leq C_i \leq C_{total},\\
        K \geq \frac{a_1}{C_1} + b_1\\
        K \geq \frac{a_2}{C_2} + b_2\\
        \vdots\\
        K \geq \frac{a_{|S|}}{C_{|S|}} + b_{|S|}.
    \end{gathered}
\end{equation}
When $C_i \geq 0$, $T^e_i$ is a convex function ($\nabla^2 T^e_i \geq 0$). Then, a convex optimization solver~\cite{mosek} can be leveraged to solve this subproblem.

\section{Experiments}
\label{sec:Experiments}
In this section, we demonstrate that our ESFL can inherently offer similar performance under the same resource limitation for all users, while significantly reducing total training time (\textit{time efficiency}). We then show the superior training performance with limited resources and limited training time (\textit{model performance}). Finally, we validate our iterative optimization approach under various system settings. 

\subsection{Experimental Setup}
For all following experiments, we evaluate the performance on image classification tasks over the common dataset CIFAR-10 and leverage VGG13, VGG16 and VGG19~\cite{simonyan2014very} framework as the neural network architecture to implement the distributed applications. We compare our ESFL with the alternatives such as FedAVG (FL), original split learning (SL), and splitfed learning (SFL).

\textbf{Dataset:}~CIFAR-10~\cite{krizhevsky2009learning}\cite{krizhevsky2010convolutional} contains $50,000$ color training images and $10,000$ testing images with $32 \times 32$ resolution in $10$ classes, with $6,000$ images per class. We assume that there are $100$ users participating in the whole training process, while only $10$ users are randomly selected to join one-round training. Under the assumption that all users' data samples are independently and identically distributed (IID), the data is shuffled and then partitioned into $100$ clients with no replacement, every user owning $500$ training samples.

\textbf{Training Configuration:}~We use a distributed machine learning framework, similar to federated learning, which has several learning hyperparameters including local learning rate $\rho_r$, where $\rho_0 = 0.01$ and $\rho_r$ is decaying as the round number $r$ increases and the constant number of local epochs $\epsilon_i = 5$. Moreover, we introduce a global learning rate $\eta = 0.5$ to control the global model updating pace. According to the experimental results, when choosing a mini-batch size of $32$, we can obtain a well trained model.

\textbf{Neural Network Architecture:}~We deploy the VGG19 network~\cite{simonyan2014very} as the training model, which primarily consists of convolutional layers (CONV), fully-connected layers (FC), and softmax layer (SoftMax). We resize the input layer of the original VGG13, VGG16, and VGG19 from $224 \times 224$ to $32 \times 32$ to fit the CIFAR-10 dataset. The mini-batch size is set to 32. We present VGG19 architecture and workload of each layer in Table~\ref{network_setting_vgg19}.

\begin{table}[!htbp]
    \centering
    \caption{VGG19 Network Architecture and Parameters}
    \begin{tabular}{c|c|c|c}
        \toprule
        \multirow{2}*{Layer} & Layer size & FP FLOPs & Activation \\
        ~ & (MBs)& (MBs) & (MBs)\\
        \midrule
        CONV1 &  0.0017 & 1.796 & 0.0655\\
        \midrule
        CONV2& 0.0369 & 37.749 & 0.0328 \\
        \midrule
        CONV3  & 0.0737 & 18.874 & 0.0328 \\
        \midrule
        CONV4 & 0.147 & 37.749 & 0.0164 \\
        \midrule
        CONV5  & 0.295 & 18.874 & 0.0164 \\
        \midrule
        CONV6 & 0.590 & 37.749 & 0.0164 \\
        \midrule
        CONV7 & 0.590 & 37.749 & 0.0164 \\
        \midrule
        CONV8 & 0.590 & 37.749 & 0.0082 \\
        \midrule
        CONV9 & 1.180 & 18.874 & 0.0082 \\
        \midrule
        CONV10 & 2.359 & 37.749 & 0.0082 \\
        \midrule
        CONV11 & 2.359 & 37.749 & 0.0082 \\
        \midrule
        CONV12 & 2.359 & 37.749 & 0.0020 \\
        \midrule
        CONV13 & 2.359 & 9.437 & 0.0020 \\
        \midrule
        CONV14 & 2.359 & 9.437 & 0.0020 \\
        \midrule
        CONV15 & 2.359 & 9.437 & 0.0020 \\
        \midrule
        CONV16 & 2.359 & 9.437 & 0.0010 \\
        \midrule
        FC1 & 102.760 & 2.097 & 4.08E-5 \\
        \midrule
        FC2 & 16.777 & 0.524 & 4.08E-5 \\
        \midrule
        FC3 & 4.096 & 0.131 & 1E-5 \\
        \midrule
        SoftMax  & $\backslash$ & $\backslash$ & $\backslash$ \\
        \bottomrule
    \end{tabular}
    \label{network_setting_vgg19}
\end{table}

\subsection{Model Performance}
\begin{figure*}[!htbp]
    \centering
    \subfigure[Testing accuracy and loss of VGG13, VGG16 and VGG19 using FL]{
        \includegraphics[width=0.22\textwidth]{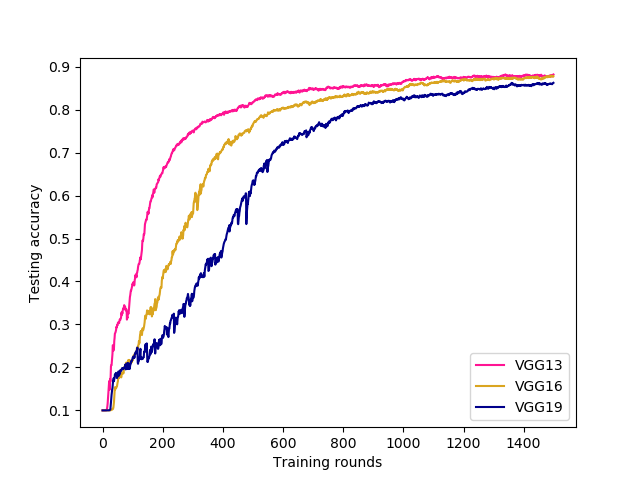}
        \includegraphics[width=0.22\textwidth]{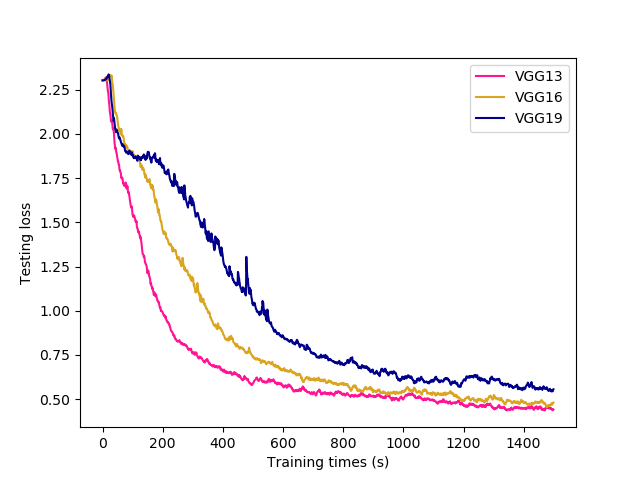}}
    \subfigure[Testing accuracy and loss of VGG13, VGG16 and VGG19 using SL]{
        \includegraphics[width=0.22\textwidth]{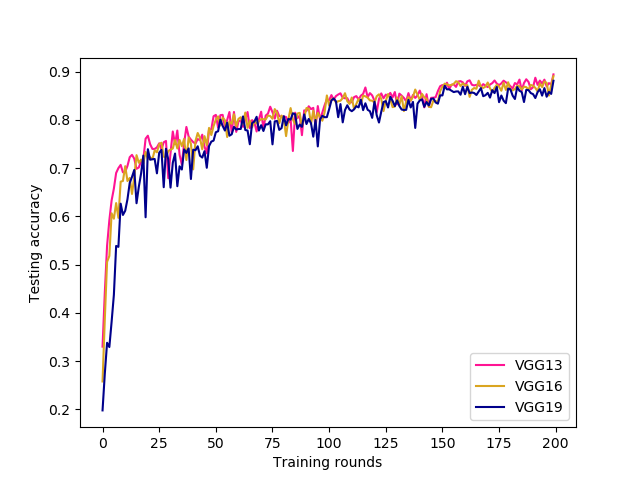}
        \includegraphics[width=0.22\textwidth]{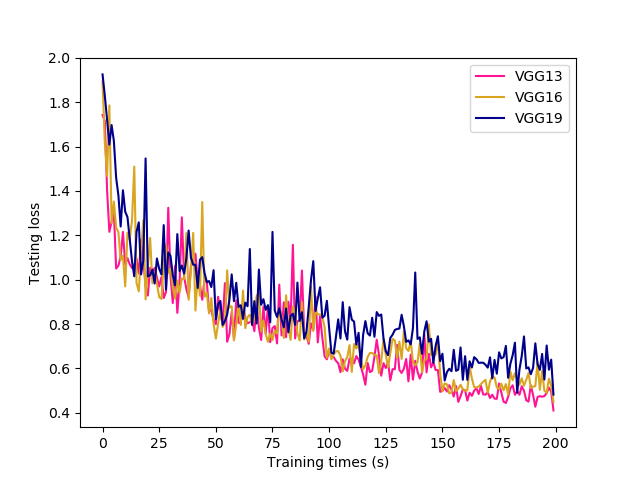}}
    \caption{Testing accuracy and loss over CIFAR-10 testing dataset for FL, SFL, ESFL and SL using three different NN (VGG13, VGG16 and VGG19). Fair comparison are guaranteed by the required training rounds to achieve the convergence threshold.}
\label{Testing_result}
\end{figure*}

One key hyperparameter in our ESFL that affect the final convergence performance such as testing accuracy and loss is the number of training rounds, since we leverage \textbf{FedAVG} for all distributed ML algorithms except SL. For a fair comparison, we set the training threshold for VGG13 to $88\%$, VGG16 to $87.5\%$ and VGG19 to $86.5\%$ testing accuracy based on the worst converged accuracy. Three distributed ML algorithms (FL, SFL, ESFL) achieve the expected convergence performance at the $1500$-th training round, and SL achieves the expected convergence performance at the $200$-th training round. The testing performance results are shown in Figure.~\ref{Testing_result}.

The rationals for choosing IID configuration rather than Non-IID in our model training process is to mitigate the impact of data distribution heterogeneities for fair training performance comparisons across different distributed ML frameworks.

\subsection{Time Efficiency}
In our simulation, at each round, the server randomly selects $10\%$ users (the \textit{selected users}) from \textit{available users} to join one-round training. Since we assume that the server possesses sufficient but limited computing resources, in this experiment, the training server is installed with an A100 GPU with $130$ teraFLOPs (TFLOPs) computing capability and $128$ GigaBytes (GBs) memory space. We compare the time efficiency of our ESFL with original federated learning (FL)~\cite{mcmahan2017communication}, split learning~\cite{gupta2018distributed} and splitfed learning (SFL)~\cite{thapa2022splitfed}.
\subsubsection{Resource Limitation}

\begin{table}[!htbp]
    \centering
    \caption{Communication and computing resource settings}
    \begin{tabular}{c|c|c}
        \toprule
        &Communication(KBps)&Computing(TFLOPs)\\
        \midrule
        BP &[$10$, $15$, $20$, $25$]&[$1.3$, $1.95$, $2.6$, $3.25$]\\
        \midrule
        PR &[$10$, $15$, $20$, $25$]&[$6.5$, $9.75$, $13$, $16.25$]\\
        \midrule
        RP &[$50$, $75$, $100$, $125$]&[$1.3$, $1.95$, $2.6$, $3.25$]\\
        \midrule
        BR &[$50$, $75$, $100$, $125$]&[$6.5$, $9.75$, $13$, $16.25$]\\
        \bottomrule
    \end{tabular}
    \label{resource_richpoor}
\end{table}

\begin{table*}[!htbp]
    \centering
    \caption{One-round training and communication time for different resource settings}
    \begin{tabular}{c|c|c|c|c|c|c|c|c|c}
        \toprule
        \multirow{2}*{}&\multirow{2}*{NNs}&\multicolumn{4}{c|}{Training and communication time(s)}&\multicolumn{4}{c}{Communication time(s)}\\ 
        \cline{3-10}
        &~& FL & SL & SFL & ESFL & FL & SL & SFL & ESFL\\
        \midrule
        \multirow{3}*{BP}&VGG13&$40.916$&$226.444$&$42.501$&\boldmath{$28.583$}&$11.182$&$123.365$&$21.968$&\boldmath{$8.726$}\\
        \cline{2-10}
        &VGG16&$52.960$&$254.787$&$48.476$&\boldmath{$31.125$}&\boldmath{$11.599$}&$115.900$&$20.284$&$12.828$\\
        \cline{2-10}
        &VGG19&$63.096$&$282.426$&$53.231$&\boldmath{$31.128$}&\boldmath{$10.200$}&$107.575$&$18.253$&$16.216$\\
        \midrule
        \multirow{3}*{PR}&VGG13&$18.351$&$148.283$&$27.775$&\boldmath{$8.245$}&$13.300$&$127.126$&$23.043$&\boldmath{$2.461$}\\
        \cline{2-10}
        &VGG16&$20.933$&$143.427$&$27.065$&\boldmath{$10.242$}&$13.401$&$114.513$&$20.616$&\boldmath{$2.858$}\\
        \cline{2-10}
        &VGG19&$23.700$&$141.961$&$27.532$&\boldmath{$12.901$}&$14.017$&$105.299$&$18.927$&\boldmath{$3.433$}\\
        \midrule
        \multirow{3}*{RP}&VGG13&$34.128$&$130.896$&$29.229$&\boldmath{$19.297$}&\boldmath{$4.238$}&$50.829$&$8.430$&$10.961$\\
        \cline{2-10}
        &VGG16&$45.713$&$156.239$&$35.828$&\boldmath{$19.657$}&\boldmath{$4.292$}&$45.390$&$7.585$&$11.135$\\
        \cline{2-10}
        &VGG19&$58.275$&$181.994$&$42.161$&\boldmath{$19.855$}&\boldmath{$4.566$}&$42.756$&$7.019$&$11.123$\\
        \midrule
        \multirow{3}*{BR}&VGG13&$10.937$&$73.141$&$14.334$&\boldmath{$6.961$}&$5.005$&$51.208$&$9.202$&\boldmath{$0.989$}\\
        \cline{2-10}
        &VGG16&$12.929$&$75.028$&$15.194$&\boldmath{$8.671$}&$4.674$&$45.801$&$8.118$&\boldmath{$2.621$}\\
        \cline{2-10}
        &VGG19&$15.968$&$78.177$&$16.220$&\boldmath{$9.470$}&$5.428$&$42.736$&$7.549$&\boldmath{$3.184$}\\
        \bottomrule
    \end{tabular}
    \label{One_training_time_re}
\end{table*}

We separate the impacts of user-side communication and computing resource limitation by simulating four resource settings shown in Table~\ref{resource_richpoor}: \textbf{Both Poor (BP)} indicates that both communication and computing resources are highly limited at EDs, \textbf{Poorcom Richcmp (PR)} indicates that communication resources are highly limited while computing resources are slightly limited (five times larger than that for the highly limited case), \textbf{Richcom Poorcmp (RP)} indicates that communication resources are slightly limited while computing resources are highly limited, and \textbf{Both Rich (BR)} indicates that both communication and computing resources are slightly limited. The selection process for each user-side resource setting is similar to Table~\ref{resource_richpoor}. For learning algorithmic implementation, we use original FL and SL, and SFL, which is similar to that for ESFL introduced in Section~\ref{sec:ESFLearning}. 

\begin{figure*}[!htbp]
    \centering
    \subfigure[Layers distributions of VGG13]{
        \includegraphics[width=0.3\textwidth]{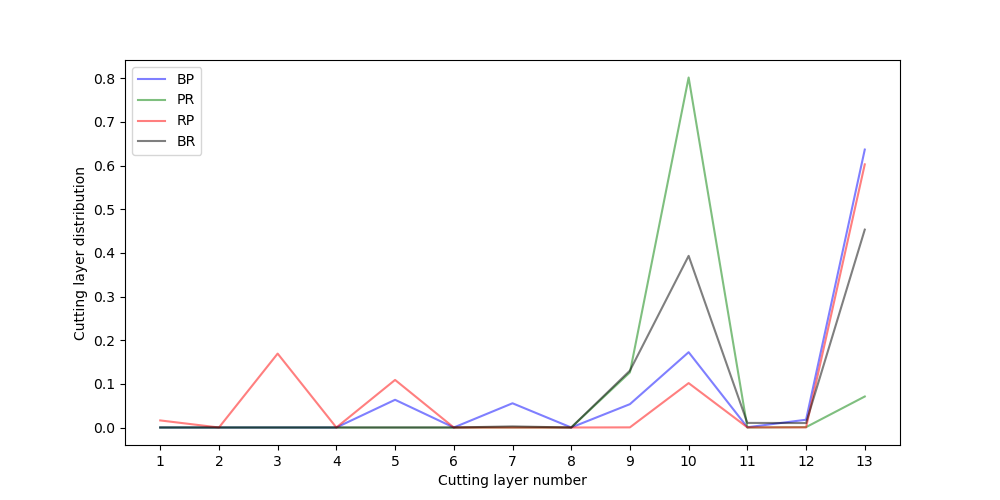}}
    \subfigure[Layers distributions of VGG16]{
        \includegraphics[width=0.3\textwidth]{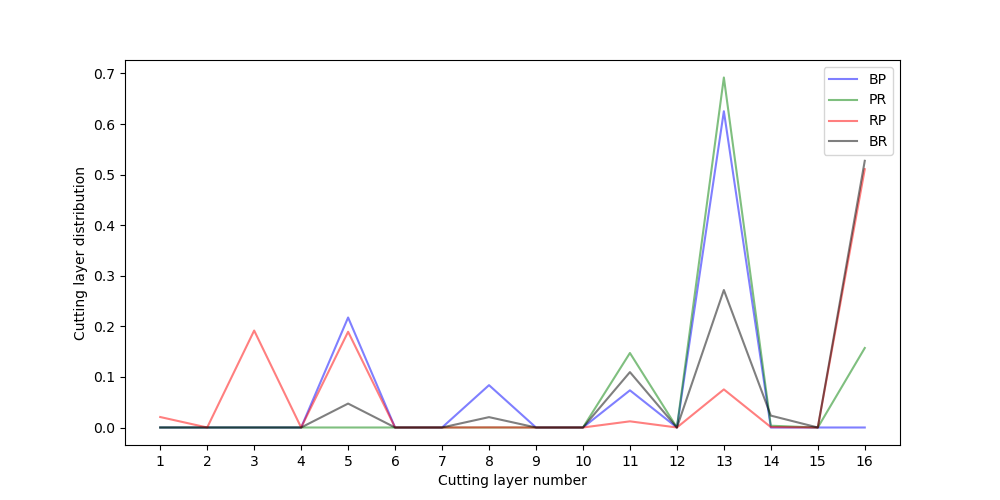}}
    \subfigure[Layers distributions of VGG19]{
        \includegraphics[width=0.3\textwidth]{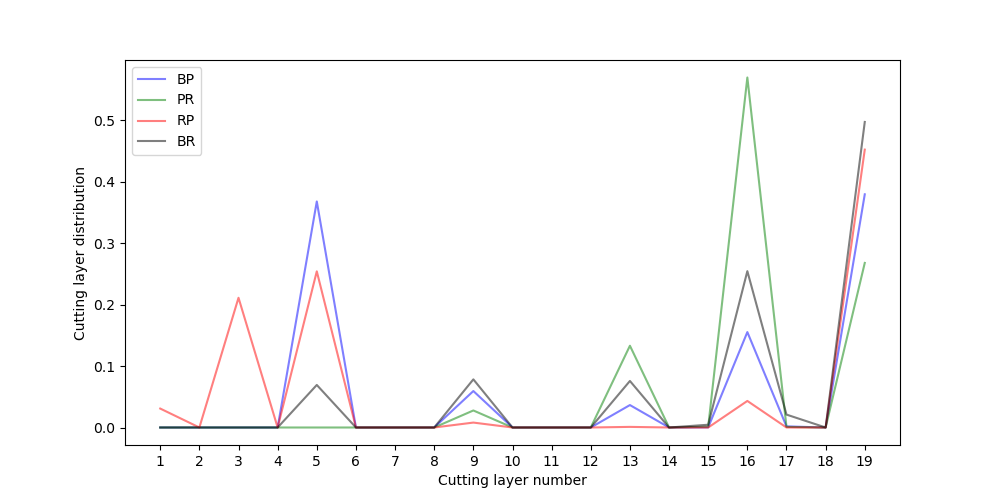}}
    \caption{Cut layer distributions (user-side workloads allocation) of three NNs under four different resource limitations using ESFL algorithm.}
\label{Distribution_re}
\end{figure*}

Table~\ref{One_training_time_re} presents the average one-round training and communication time and one-round communication time for different NNs under different resource scenarios using FL, SL, SFL and efficient split federated learning ESFL, respectively. Figure.~\ref{Distribution_re} shows the allocation results of user-side training workload represented as cut layer distributions. The cut layer distribution represents the empirical probability of selecting layer $l$ for user $i$ in the total training rounds, which is $P_{i,l} = \sum^R_r \frac{x_{i,r}^l}{R}$, where $x_{i,r}^l$ is the cutting layer decision showing in equation~\ref{Epoch_time} and $\sum_l P_{i,l} = 1$. The cut layer distribution combining with the amounts of user-side data indicates the allocated user-side computing and communication workload. Therefore, from cut layer distributions, as the user-side resource becomes more sufficient, our ESFL applies more identical cut layer distributions strategies for all NNs. For the results in those two tables, our ESFL algorithm significantly reduces one-round training and communication time under all circumstances. These advantages often stem from the dynamics between user-side computing and communication resource. In scenarios where local computing resource are poor whereas communication resource are rich (\textbf{RP}), ESFL remains fewer layers of user-side models by leveraging more on server-side computing power. Conversely, under the \textbf{PR} scenario, ESFL mitigates these limitations by remaining more layers of user-side model to rely less on communication. Comparing one-round training and communication time of SFL and FL, an intriguing phenomenon emerges. Although SFL leverages the server-side resource to accelerate training, improper user-side workload allocation (model separation) and server-side resource allocation lead to decreased time efficiency. 

\begin{table*}[!htbp]
    \centering
    \caption{Total training and communication time for different resource settings}
    \begin{tabular}{c|c|c|c|c|c|c|c|c|c}
        \toprule
        \multirow{2}*{}&\multirow{2}*{NNs}&\multicolumn{4}{c|}{Training and communication time(s)}&\multicolumn{4}{c}{Communication time(s)}\\ 
        \cline{3-10}
        &~& FL & SL & SFL & ESFL & FL & SL & SFL & ESFL \\
        \midrule
        \multirow{3}*{BP}&VGG13&$61,374$&$45,288$&$63,751$&\boldmath{$42,874$}&$16,773$&$24,673$&$32,952$&\boldmath{$13,089$}\\
        \cline{2-10}
        &VGG16&$79,440$&$50,957$&$72,714$&\boldmath{$46,687$}&\boldmath{$17,398$}&$23,180$&$30,426$&$19,242$\\
        \cline{2-10}
        &VGG19&$94,644$&$56,485$&$79,725$&\boldmath{$46,692$}&\boldmath{$15,300$}&$21,515$&$27,379$&$24,324$\\
        \midrule
        \multirow{3}*{PR}&VGG13&$27,526$&$29,656$&$41,662$&\boldmath{$12,367$}&$19,950$&$25,425$&$34,564$&\boldmath{$3,691$}\\
        \cline{2-10}
        &VGG16&$31,399$&$28,685$&$41,298$&\boldmath{$15,363$}&$20,101$&$22,902$&$30,924$&\boldmath{$4,287$}\\
        \cline{2-10}
        &VGG19&$35,550$&$28,392$&$41,298$&\boldmath{$19,351$}&$21,025$&$21,059$&$28,390$&\boldmath{$5,149$}\\
        \midrule
        \multirow{3}*{RP}&VGG13&$51,192$&\boldmath{$26,179$}&$43,843$&$28,945$&\boldmath{$6,357$}&$10,165$&$12,645$&$16,441$\\
        \cline{2-10}
        &VGG16&$68,569$&$31,247$&$53,742$&\boldmath{$29,485$}&\boldmath{$6,438$}&$9,078$&$11,377$&$16,702$\\
        \cline{2-10}
        &VGG19&$87,412$&$36,398$&$63,241$&\boldmath{$29,782$}&\boldmath{$6,849$}&$64,134$&$10,528$&$16,684$\\
        \midrule
        \multirow{3}*{BR}&VGG13&$16,405$&$14,628$&$21,501$&\boldmath{$10,441$}&$7,507$&$10,241$&$13,803$&\boldmath{$1,483$}\\
        \cline{2-10}
        &VGG16&$19,393$&$15,005$&$22,791$&\boldmath{$13,006$}&$7,011$&$9,160$&$12,177$&\boldmath{$3,931$}\\
        \cline{2-10}
        &VGG19&$23,952$&$15,635$&$24,330$&\boldmath{$14,205$}&$8,142$&$8,547$&$11,323$&\boldmath{$4,776$}\\
        \bottomrule
    \end{tabular}
    \label{Total_training_time_re}
\end{table*}

Table~\ref{Total_training_time_re} indicates the total training and communication time for four ML algorithms to achieve the expected convergence performance. In the context of contrasting FL and SFL, it is imperative to acknowledge that although SFL, similar to ESFL, utilizes server-side computing resource, its overall performance is significantly influenced by the harmonization of user-side workload and server-side resource allocation. The lack of effective resource allocation strategies can result in inferior performance in SFL when compared to FL. Especially in \textbf{RP} and \textbf{BR}, when communication resource is notably limited in EDs, FL outperforms SFL. This performance discrepancy is attributed to the lack of effective allocating strategies, which results in inferior performance in SFL when compared to FL. Nevertheless, our ESFL exhibits a significant increase in efficiency compared to both FL and SFL across all tested scenarios. This provides the evidence that a well-conceived strategy for workload and resource allocation can markedly enhance the efficiency of the whole training process. While the model used here and other ML algorithms are not the state-of-the-art for this task, it does provide sufficient evidence to show that our ESFL can significantly reduce training latency and improve training efficiency by considering user-side resource heterogeneity.

\subsubsection{Resource Heterogeneity}

\begin{table}[!htbp]
    \centering
    \caption{Communication and computing resource settings}
    \begin{tabular}{c|c|c}
        \toprule
        & Communication(KBps) & Computing(TFLOPs)\\
        \midrule
        SH &[$10$, $15$, $20$, $25$]&[$1.3$, $1.95$, $2.6$, $3.25$]\\
        \midrule
        SL &[$10$, $15$, $20$, $25$]&[$0.65$, $1.3$, $2.6$, $4.55$]\\
        \midrule
        LS &[$5$, $10$, $20$, $35$]&[$1.3$, $1.95$, $2.6$, $3.25$]\\
        \midrule
        LH &[$5$, $10$, $20$, $35$]&[$0.65$, $1.3$, $2.6$, $4.55$]\\
        \bottomrule
    \end{tabular}
    \label{resource_setting}
\end{table}

\begin{table*}[!htbp]
    \centering
    \caption{One-round training and communication time for different heterogeneities}
    \begin{tabular}{c|c|c|c|c|c|c|c|c|c}
        \toprule
        \multirow{2}*{}&\multirow{2}*{NNs}&\multicolumn{4}{c|}{Training and communication time(s)}&\multicolumn{4}{c}{Communication time(s)}\\ 
        \cline{3-10}
        &~& FL & SL & SFL & ESFL & FL & SL & SFL & ESFL\\
        \midrule
        \multirow{3}*{SH}&VGG13&$50.706$&$362.412$&$66.277$&\boldmath{$34.043$}&$22.853$&$257.278$&$45.766$&\boldmath{$6.170$}\\
        \cline{2-10}
        &VGG16&$64.136$&$362.872$&$67.640$&\boldmath{$40.658$}&$24.544$&$225.720$&$40.473$&\boldmath{$11.093$}\\
        \cline{2-10}
        &VGG19&$76.288$&$381.228$&$69.879$&\boldmath{$43.074$}&$24.063$&$210.248$&$35.558$&\boldmath{$22.941$}\\
        \midrule
        \multirow{3}*{SL}&VGG13&$78.595$&$418.746$&$83.431$&\boldmath{$55.912$}&$19.913$&$251.753$&$44.067$&\boldmath{$17.867$}\\
        \cline{2-10}
        &VGG16&$103.394$&$457.030$&$92.968$&\boldmath{$58.576$}&\boldmath{$21.957$}&$229.795$&$39.047$&$26.674$\\
        \cline{2-10}
        &VGG19&$126.801$&$502.072$&$105.241$&\boldmath{$61.022$}&\boldmath{$22.728$}&$214.884$&$36.744$&$31.815$\\
        \midrule
        \multirow{3}*{LS}&VGG13&$79.478$&$518.936$&$109.171$&\boldmath{$37.625$}&$53.200$&$415.204$&$90.510$&\boldmath{$9.832$}\\
        \cline{2-10}
        &VGG16&$87.448$&$512.143$&$109.142$&\boldmath{$48.904$}&$51.658$&$369.021$&$82.441$&\boldmath{$10.685$}\\
        \cline{2-10}
        &VGG19&$95.622$&$522.882$&$106.500$&\boldmath{$54.211$}&$49.811$&$347.786$&$74.140$&\boldmath{$19.720$}\\
        \midrule
        \multirow{3}*{LH}&VGG13&$91.815$&$565.541$&$123.469$&\boldmath{$61.865$}&$41.606$&$399.900$&$86.496$&\boldmath{$14.211$}\\
        \cline{2-10}
        &VGG16&$117.439$&$608.637$&$130.643$&\boldmath{$73.708$}&$40.871$&$386.282$&$80.686$&\boldmath{$17.929$}\\
        \cline{2-10}
        &VGG19&$145.186$&$601.846$&$130.470$&\boldmath{$75.520$}&$44.473$&$323.242$&$66.901$&\boldmath{$42.704$}\\
        \bottomrule
    \end{tabular}
    \label{One_training_time_he}
\end{table*}

\begin{figure*}[!htbp]
    \centering
    \subfigure[Layers distributions of VGG13]{
        \includegraphics[width=0.3\textwidth]{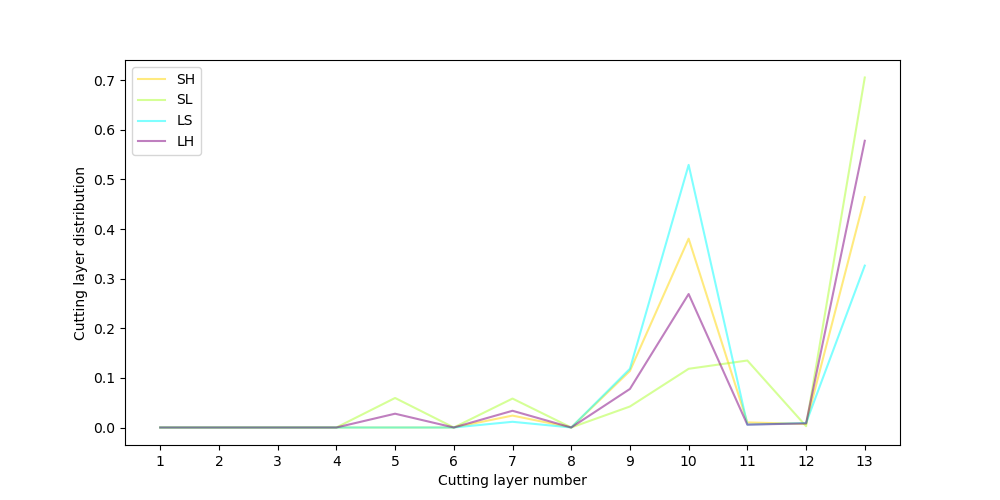}}
    \subfigure[Layers distributions of VGG16]{
        \includegraphics[width=0.3\textwidth]{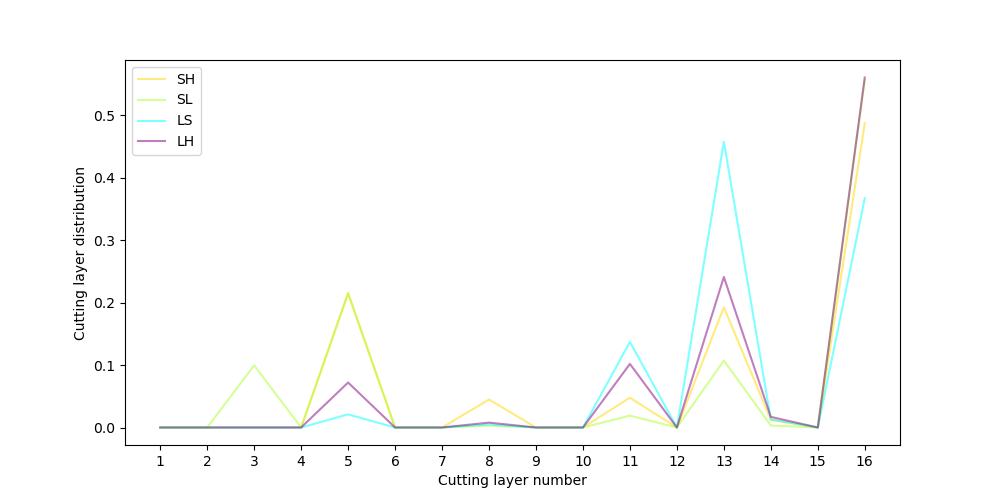}}
    \subfigure[Layers distributions of VGG19]{
        \includegraphics[width=0.3\textwidth]{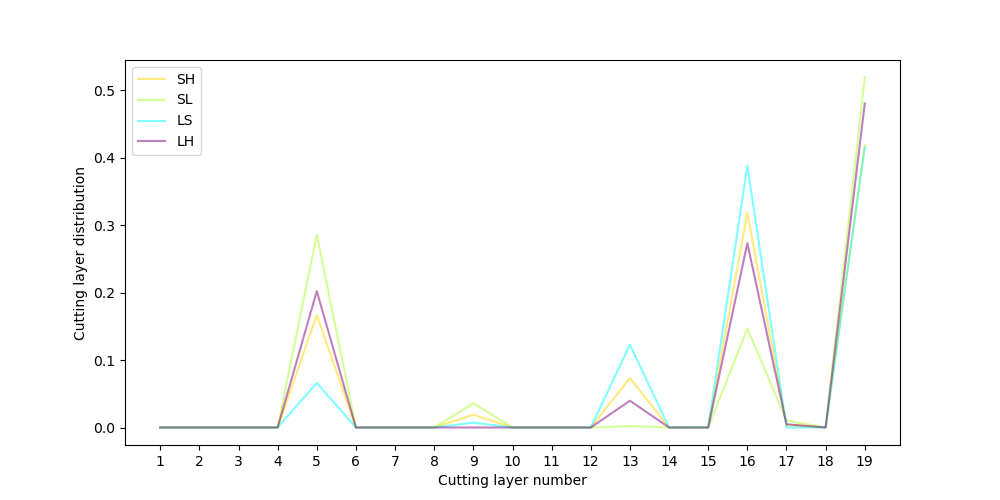}}
    \caption{Cut layer distributions (user-side workloads allocation) of three NNs under four different resource heterogeneities using ESFL algorithm.}
\label{Distribution_he}
\end{figure*}

In this section, we assume there exist four different communication and computing resource settings to evaluate the efficiency of our ESFL algorithm under different heterogeneous scenarios. The detail of our resource simulation setting is shown in Table~\ref{resource_setting}, where in every training round, the available communication and computing conditions of the \textit{selected users} are randomly chosen from resource options in one scenario. In particular, considering the fairness, the average resource amounts in different scenarios are equal, and only resource distributions are dissimilar to simulate different resource heterogeneity. Four heterogeneous scenarios are considered: \textbf{Small Heterogeneity (SH)}, implying that both communication and computing resource heterogeneity at EDs is small, \textbf{Smallcom Largecmp (SL)}, implying that communication heterogeneity is small while computing resource is large, \textbf{Largecom Smallcmp (LS)}, implying that communication heterogeneity is large while computing resource is small, and \textbf{Large Heterogeneity (LH)}, implying that both communication and computing resource heterogeneities are large. For instance, in \textbf{SH}, each user will randomly choose one communication condition from [$10$, $15$, $20$, $25$] kiloBytes (KBps) and one computing condition from [$1.3$, $1.95$, $2.6$, $3.25$] TFLOPs as their available resources, and the server will base on the resource information of the selected users to allocate appropriate server-side computing resource and make user-side cutting layer decision for all selected users. To simulate the training workload heterogeneity, we assume that all \textit{available users} have heterogeneous but constant amounts of data samples, which are chosen from [$200$, $400$, $600$, $800$].

Table~\ref{One_training_time_he} presents the average one-round training and communication time and one-round communication time, for different NNs under different resource scenarios using FL, SL, SFL and ESFL, respectively. For the results in Table~\ref{One_training_time_he}, our ESFL algorithm significantly reduces one-round training and communication time under all circumstances and is least affected by resource heterogeneity, where both communication and computing heterogeneity seriously impact the training efficiencies of the other three ML algorithms. It is noteworthy that as the distribution of resources approaches a state of greater uniformity (\textbf{SH}), the gap of time efficiency between SFL and ESFL is decreased. Conversely, with an increase in resource heterogeneity (\textbf{LH}), the performance differential between ESFL and SFL widens notably, which highlights the robustness of ESFL in diverse resource environments. For instance, the training latency of SFL in \textbf{LH} is increased by nearly two times compared with that in \textbf{SH} while training latencies of ESFL are nearly the same in all scenarios. The user-side training workload allocation results (cut layer distributions) is shown in Figure.~\ref{Distribution_he}. From this simulation result, ESFL algorithm demonstrates a trend implementing more uniform cut layer distributions across all NNs, under more identical resource distributions.

\begin{table*}[!htbp]
    \centering
    \caption{Total training time to achieve the convergence performance for different heterogeneities}
    \begin{tabular}{c|c|c|c|c|c|c|c|c|c}
        \toprule
        \multirow{2}*{}&\multirow{2}*{NNs}&\multicolumn{4}{c|}{Training and communication time(s)}&\multicolumn{4}{c}{Communication time(s)}\\ 
        \cline{3-10}
        &~& FL & SL & SFL & ESFL & FL & SL & SFL & ESFL \\
        \midrule
        \multirow{3}*{SH}&VGG13&$76,059$&$72,482$&$99,415$&\boldmath{$51,064$}&$34,279$&$51,455$&$68,649$&\boldmath{$9,255$}\\
        \cline{2-10}
        &VGG16&$96,204$&$72,574$&$101,460$&\boldmath{$60,987$}&$36,816$&$45,144$&$60,709$&\boldmath{$16,639$}\\
        \cline{2-10}
        &VGG19&$114,432$&$72,574$&$104,818$&\boldmath{$64,611$}&$36,816$&$42,049$&$53,337$&\boldmath{$34,411$}\\
        \midrule
        \multirow{3}*{SL}&VGG13&$117,892$&\boldmath{$83,749$}&$125,146$&$83,868$&$29,869$&$50,350$&$66,100$&\boldmath{$26,800$}\\
        \cline{2-10}
        &VGG16&$155,091$&$91,406$&$139,452$&\boldmath{$84,217$}&\boldmath{$32,935$}&$45,959$&$58,570$&$40,011$\\
        \cline{2-10}
        &VGG19&$190,201$&$100,414$&$157,861$&\boldmath{$91,533$}&\boldmath{$34,092$}&$42,976$&$55,122$&$47,722$\\
        \midrule
        \multirow{3}*{LS}&VGG13&$119,217$&$103,787$&$163,756$&\boldmath{$56,437$}&$79,800$&$83,040$&$135,765$&\boldmath{$14,748$}\\
        \cline{2-10}
        &VGG16&$131,232$&$102,428$&$163,713$&\boldmath{$73,356$}&$77,487$&$73,804$&$123,661$&\boldmath{$16,027$}\\
        \cline{2-10}
        &VGG19&$143,433$&$104,576$&$159,750$&\boldmath{$81,316$}&$74,716$&$69,557$&$111,210$&\boldmath{$29,580$}\\
        \midrule
        \multirow{3}*{LH}&VGG13&$137,722$&$113,108$&$185,203$&\boldmath{$92,797$}&$62,409$&$79,980$&$129,744$&\boldmath{$21,316$}\\
        \cline{2-10}
        &VGG16&$176,158$&$121,727$&$195,964$&\boldmath{$110,562$}&$61,306$&$77,256$&$121,029$&\boldmath{$26,893$}\\
        \cline{2-10}
        &VGG19&$217,779$&$120,369$&$195,705$&\boldmath{$113,280$}&$66,709$&$64,648$&$100,351$&\boldmath{$64,056$}\\
        \bottomrule
    \end{tabular}
    \label{Total_training_time_he}
\end{table*}

Table~\ref{Total_training_time_he} presents the total training and communication time under different \textbf{RH}. It can be seen from the results that our ESFL method is significantly more efficient compared with the original FL, SL, and SFL in most scenarios, only except VGG13 in \textbf{SL}. Comparing communication time of FL in SL and LS, the performance of FL is markedly impacted by the \textbf{RH}. However, ESFL capitalizes on these heterogeneities through joint workload and resource allocation: in the environment with low  communication heterogeneity (\textbf{SL}), it increases the user-side communication workload while decreases computing workload; in the environment with high communication heterogeneity (\textbf{LS}), it conversely allocates more computing workload to EDs. Therefore, our approach optimizes the time efficiency across diverse scenarios by adaptively leveraging \textbf{RH}.

\subsection{Resource Allocation Convergence Analysis}

\begin{figure*}
    \centering
    \subfigure[Average one-round training time in \textbf{BP}]{
        \includegraphics[width=0.45\textwidth]{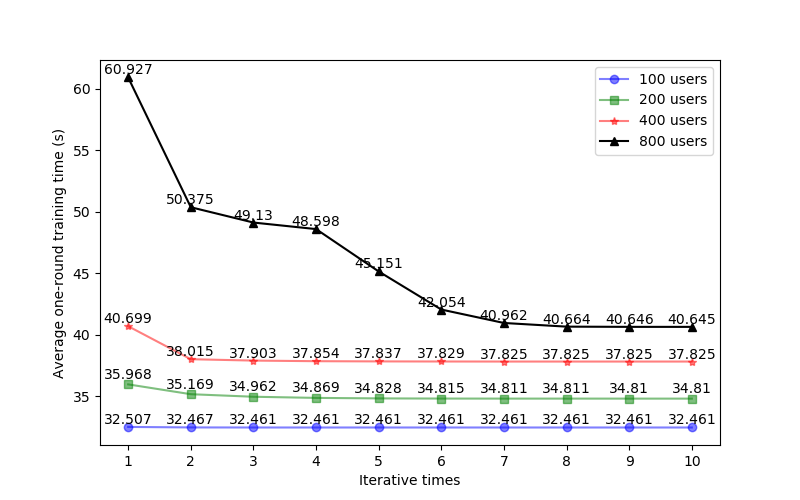}}
    \subfigure[Average one-round training time in \textbf{PR}]{
        \includegraphics[width=0.45\textwidth]{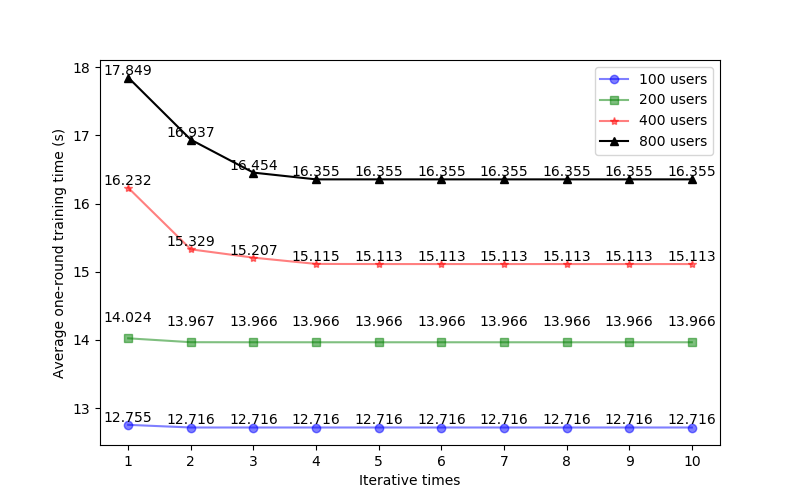}}
    \quad
    \subfigure[Average one-round training time in \textbf{RP}]{
        \includegraphics[width=0.45\textwidth]{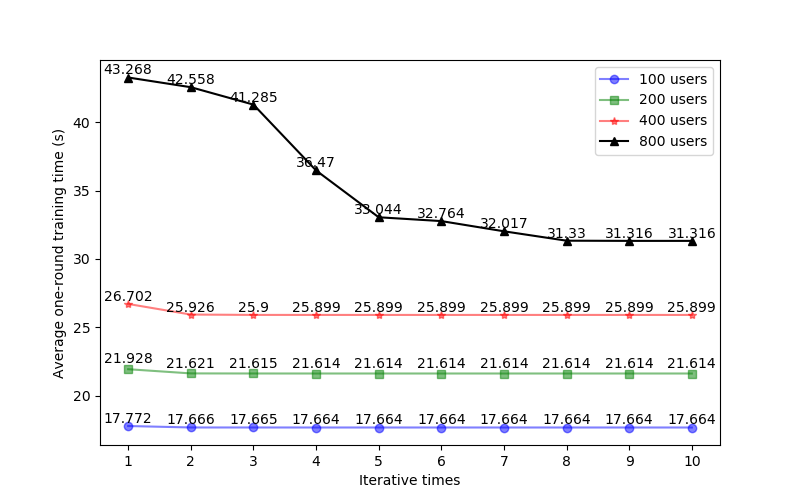}}
    \subfigure[Average one-round training time in \textbf{BR}]{
        \includegraphics[width=0.45\textwidth]{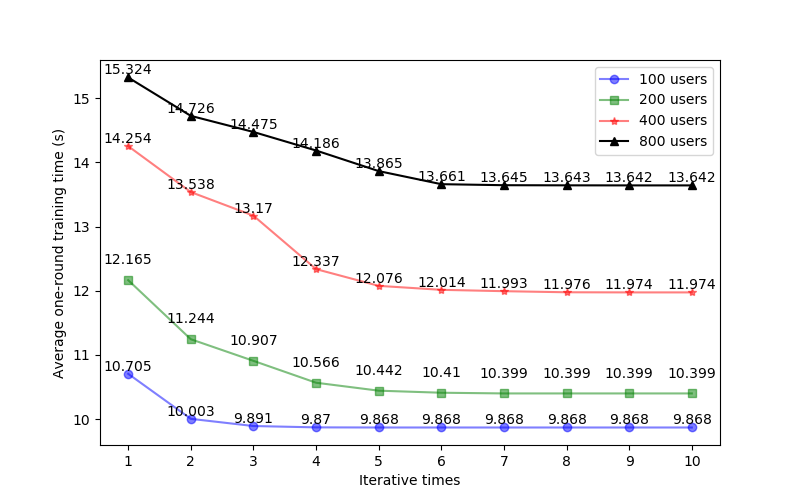}}
    \caption{Average one-round training time using iterative optimization of VGG19 in four different resource scenarios shown in Table~\ref{resource_richpoor}.}
    \label{Convergence_analysis}
\end{figure*} 

From Section~\ref{sec:Optimization}, the joint resource allocation and model splitting problem has been decomposed into two subproblems due to the time complexity, and we leverage an alternative optimization approach to solve those subproblems. To evaluate the convergence of our alternative method, we simulate iterative results at four users' scales in four resource scenarios (Table~\ref{resource_richpoor}), \textbf{100 users}, \textbf{200 users}, \textbf{400 users}, and \textbf{800 users}. The simulation results in Figure.~\ref{Convergence_analysis} shows that even for the largest user scale (\textbf{800 users}) in all different resource scenarios, our alternative approach only needs $9$ iterations to achieve convergence, while when the user scales are small, it only requires a few iterations to achieve convergence. Comparing average one-round training time across different scenarios, our method significantly enhances training efficiency, particularly in the resource-constrained scenario (\textbf{BP}).

\section{Conclusion}
\label{sec:conclusion}
In this paper, we have designed ESFL, a novel distributed training approach that tackles the resource heterogeneity inherent in both federated learning and split learning. Unlike previous methods in addressing data heterogeneity in FL, we have provided a new perspective by allocating appropriate server-side resources and user-side workload to effectively address the straggler problem in the synchronous FL framework. By evaluating the training efficiency for different ML algorithms under different heterogeneous scenarios, we have performed extensive analysis and demonstrated the superiority of our proposed ESFL.

\if\hasBibliography1
\bibliographystyle{IEEEtran}
\bibliography{IEEEabrv,bibliography/zhu}
\fi

\end{document}